\documentclass[10pt,twocolumn,letterpaper]{article}

\usepackage{cvpr}              
\definecolor{cvprblue}{rgb}{0.21,0.49,0.74}
\usepackage[pagebackref,breaklinks,colorlinks,allcolors=cvprblue]{hyperref}
\usepackage{amsmath}
\usepackage{multirow}
\usepackage{xspace}
\usepackage{hyperref}
\usepackage[table]{xcolor}
\usepackage{float}
\usepackage{amssymb}
\usepackage{amsthm}

\usepackage{subcaption}
\usepackage{enumitem}
\newcommand{\minisection}[1]{\vspace{0.02in}\noindent{\bf #1}}
\newcommand{\encoder}{\mathcal{E}}
\newcommand{\ourmethod}{\textit{LumiCtrl}\xspace}
\newcommand{\decoder}{\mathcal{D}}
\newcommand{\inputimage}{\mathcal{I}}
\newcommand{\expec}{\mathbb{E}}
\newcommand{\model}{\epsilon_\theta}

\newcommand{\textprompt}{\mathcal{P}}
\newcommand{\textembedding}{\mathcal{C}}

\newcommand{\concepttoken}{\ensuremath{\mathit{[v]}}}
\newcommand{\illumtoken}{\ensuremath{[\mathit{c}^{*}_i]}}
\newcommand{\tokenset}{\mathcal{V}}
\newcommand{\mask}{\mathcal{M}}
\newcommand{\quotes}[1]{``#1''}
\newcommand{\quotesingle}[1]{`#1'}

\definecolor{rank1}{HTML}{fff1e6}
\definecolor{rank2}{HTML}{dfe7fd}
\definecolor{rank3}{HTML}{eaf4f4}

\def\paperID{20} 
\def\confName{CVPR}
\def\confYear{2026}

\title{\ourmethod: Learning Illuminant Prompts for Lighting Control in  Personalized Text-to-Image Models}

\author{
Muhammad Atif Butt$^{1,3}$, Kai Wang$^{1,2,4}$\thanks{Corresponding author.},  Javier Vazquez-Corral$^{1,3}$, Joost Van De Weijer$^{1,3}$ \\
$^{1}$ Computer Vision Center, Spain $^{2}$ City University of Hong Kong\\
$^{3}$ Computer Sciences Department, Universitat Autònoma de Barcelona, Spain \\
$^{4}$ Program of Computer Science, City University of Hong Kong (Dongguan) \\
\tt\small\{mabutt,  kwang, jvazquez, joost\}@cvc.uab.es \\
}

\begin{document}
\maketitle

\begin{abstract}
Text-to-image (T2I) models have demonstrated remarkable progress in creative image generation, yet they still lack precise control over scene illuminants which is a crucial factor for content designers to manipulate visual aesthetics of generated images. In this paper, we present an illuminant personalization method named \ourmethod that learns illuminant prompt given single image of the object. \ourmethod consists of three components: given an image of the object, our method apply (a) physics-based illuminant augmentation along with Planckian locus to create fine-tuning variants under standard illuminants; (b) Edge-Guided Prompt Disentanglement using frozen ControlNet to ensure prompts focus on illumination, not the structure; and (c) a Masked Reconstruction Loss that focuses learning on foreground object while allowing background to adapt contextually—enabling what we call Contextual Light Adaptation. We qualitatively and quantitatively compare \ourmethod against other T2I customization methods. The results show that \ourmethod achieves significantly better illuminant fidelity, aesthetic quality, and scene coherence compared to existing baselines. A human preference study further confirms the strong user preference for \ourmethod generations.
\end{abstract}
\vspace{-9mm}
\section{Introduction}
\label{intro}
With the advent of diffusion models, text-to-image (T2I) generation has witnessed unprecedented progress, enabling the synthesis of highly realistic images from natural language text prompts ~\citep{huang2025t2i}. These models generate images by gradually denoising a randomly initialized latent representation, conditioned on a text prompt, through a learned reverse diffusion process. Among the most influential architectures in this domain is Stable Diffusion~\citep{esser2024scaling}, a latent diffusion model that leverages a pre-trained autoencoder to operate in a compressed latent space, reducing computational costs while maintaining image fidelity. Stable Diffusion has demonstrated state-of-the-art performance in aligning generated imagery with text prompts, supporting a wide range of applications ranging from creative design to digital art.

\begin{figure}
  \centering
  \includegraphics[width=0.95\linewidth]{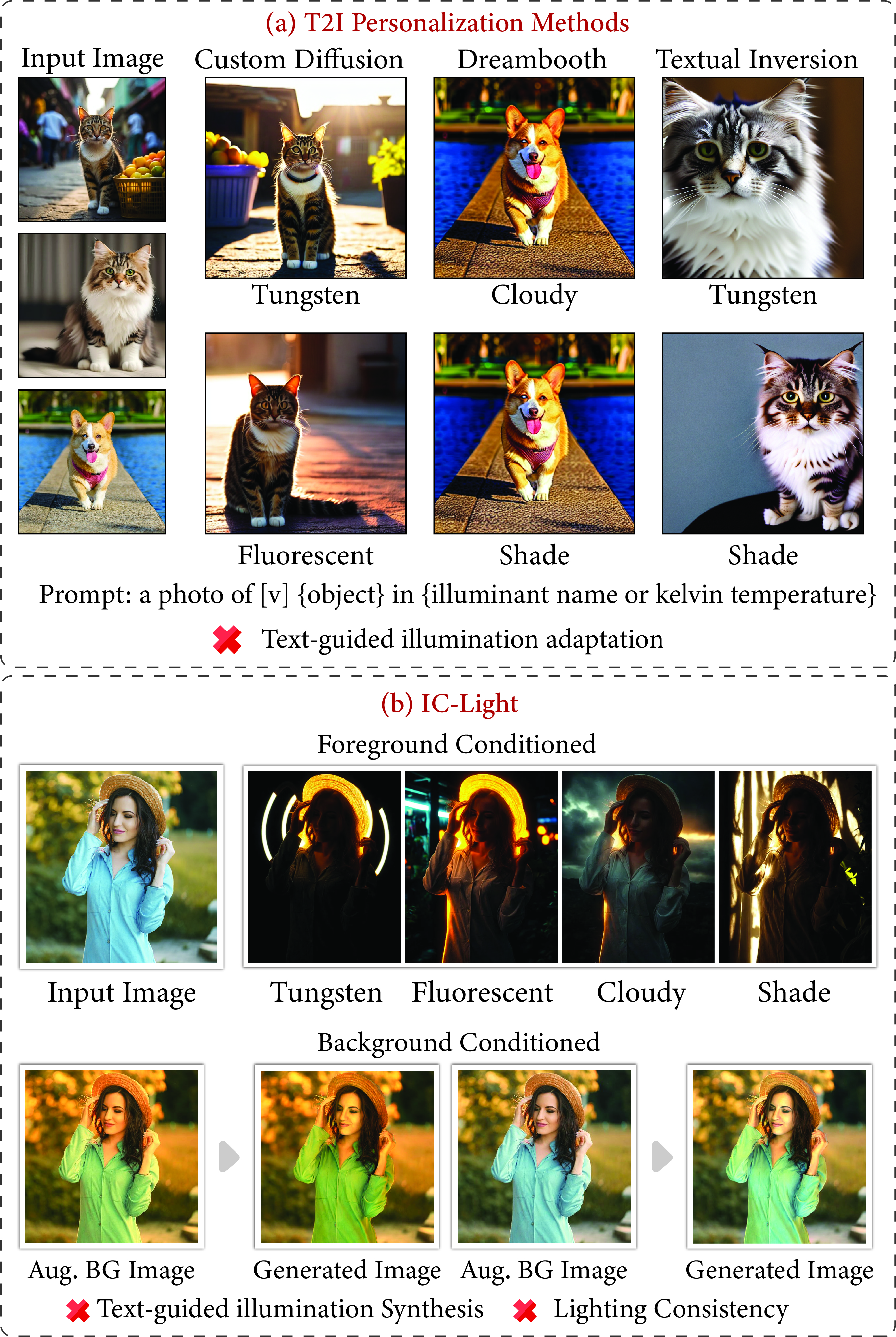}
  \vspace{-2mm}
  \caption{Limitations of existing T2I methods. (a) T2I personalization methods predominantly preserve lighting from training examples, failing to generate concepts under various illuminants. (b)  IC-Light with foreground condition fails to produce text-guided illumination and preserve background, while background conditioning with augmented image produces inconsistent lighting.
  }

  \label{fig:baseline_comparison}
  \vspace{-5mm}
\end{figure}

One of the crucial parameters that content designers like to control in images is the scene illuminant; altering this can greatly impact the mood, atmosphere, and overall attractiveness of the generated image~\citep{ge2025expressive}. Popular photo editing software like Photoshop allows users to manipulate scene illuminant using a set of five default options, including \emph{daylight}, \emph{tungsten}, \emph{fluorescent}, \emph{cloudy} and \emph{shade}. This precise illuminant control is also highly desirable for image generation\footnote{Note that in image generation we are going in the reverse direction than in post-camera softwares: We introduce an illuminant, not remove it.}.
However, current diffusion models do not allow for this control, as evident in Fig.~\ref{fig:baseline_comparison}. These models struggle in understanding the standard illuminants and their corresponding numerical representations---not because they lack visual knowledge of illumination, but because the standard illuminant terms are not grounded in their text representations. Interestingly, we find that the key reason for this failure lies in the text encoder itself: as we demonstrate in Section~\ref{subsec:embedding_analysis}, embeddings for standard illuminant terms like “tungsten” or “6500K” are not semantically grounded---they neither cluster with general lighting concepts (e.g., “warm”, “cool”) nor with their own Kelvin or named counterparts. Instead, Kelvin temperature values embed near generic numerals, revealing a fundamental disconnect between linguistic labels and photometric meaning.

Given the lack of explicit illuminant control in diffusion-based generation models, several post-hoc methods~\citep{zhang2025scaling,zeng2024rgbx,kocsis2023intrinsic,zeng2024dilightnet} are employed for image relighting which we refer as Image-Space Illumination Control (ISIC) methods. These methods take an explicit illuminant color or illuminant environment map and an image as input and modify its appearance to simulate a change in illuminant color. In contrast to our Prompt-Space Illumination Modeling (PSIM) approach, ISIC methods require a separate relighting model or software, whereas our method enables direct illuminant control within generative prompt itself. Among ISIC methods, most common technique applies a global illuminant shift assuming a single uniform light source—typically implemented as a channel-wise scaling following von Kries model~\citep{finlayson1994color}. We refer to this strategy as Flat Light Adaptation to emphasize its spatially uniform nature. This class also includes more advanced approaches based on intrinsic image decomposition~\citep{zeng2024rgbx} and hybrid methods such as IC-Light~\citep{zhang2025scaling}, which combine limited prompt-level modulation on real-images. However, this method fails to preserve spatial and semantic structure of given image, often introduces artifacts or inconsistent scene geometry, as shown in Fig.~\ref{fig:baseline_comparison}. 


In this paper, 
we propose first Prompt-Space Illumination Modeling method: our method allows for direct illumination control in the prompt that is used to generate the image. Our method builds upon recent advances in T2I personalization techniques such as Textual Inversion~\citep{textual_inversion}, DreamBooth~\citep{ruiz2022dreambooth}, and Custom Diffusion~\citep{kumari2023multi} which allow users to personalize new concepts (e.g., specific objects or pets) into pre-trained diffusion models by learning new textual embeddings. However, these models often entangle lighting information from training images, which consequently fails to generate concepts under text-guided illuminants, as shown in Fig.~\ref{fig:baseline_comparison}.
Here, we extend these personalization methods to learn \emph{illuminant prompts} for precise illuminant control in T2I generation with \ourmethod. 
We aim to leverage prior knowledge of scene illuminants embedded in large-scale diffusion models to introduce a more realistic \emph{Contextual Light Adaptation}. To disentangle illuminant from image content, we introduce two technological contributions. We propose \emph{edge-guided prompt disentanglement}, which exploits ControlNet~\citep{zhang2023controlnet} to allow prompt training to focus only on illuminant color changes, and discard structural information of training images. In addition, we introduce a new \textit{Masked Reconstruction Loss} focusing on learning illuminant color of foreground object, given a user-provided mask. This allows us to achieve \textit{Contextual Light Adaptation}, where \ourmethod dynamically adjusts background color, taking background content into account.

In summary, the main contributions are as follows:
\begin{enumerate}
    \item We discover a fundamental semantic gap: Standard illuminant terms (e.g., tungsten, 6500K) are not grounded in text encoder’s embedding space, explaining why naive illuminant prompting fails in T2I generative models.
    \item We are first to perform \textit{illuminant prompts learning}. We propose \ourmethod to learn \emph{illuminant prompts} that allow for precise \textit{illuminant control} of generated images. For illuminant learning, we propose to apply Flat Light Adaptation to obtain training images to be used as illuminance reference. We introduce \emph{Edge-Guided Prompt Disentanglement} to mitigate spurious detail leakage.
    \item To improve prompt learning, we propose a Masked Reconstruction Loss focusing on foreground objects to enable \emph{Contextual Light Adaptation} to background content.
    \item \ourmethod outperforms other prompt-space illumination modeling methods that are broadly categorized into T2I personalization, and appearance editing methods on various quantitative results and a user study.
\end{enumerate}

\vspace{-2mm}
\section{Related Work}
\label{sec:rel_work}

\subsection{Text-to-Image Diffusion Models.}
Recently, an unprecedented progress has been made in T2I generation. T2I diffusion models~\citep{esser2024scaling} emerged as more efficient models surpassing GANs~\citep{kang2023scaling, sauer2021projected}, and autoregressive~\citep{yu2022scaling, esser2021taming} models in T2I generation~\citep{xing2025training}. Diffusion models are probabilistic generative models that aim to learn data distribution through iterative denoising from Gaussian distribution. To improve controllability, these models can be conditioned on class, image, or text prompts~\citep{huang2025diffusion}. We build \ourmethod over a latent diffusion model that learns data distribution within low-dimensional space, making the training process computationally efficient while reducing inference time without compromising generation quality.

\subsection{T2I Personalization.}
T2I diffusion model personalization learns a new concept given few images and bind it to a new text token. T2I personalization methods~\citep{gu2024mixofshow, li2023photomaker} can synthesize different variations of a newly learned concept by composing text prompts. Textual Inversion~\citep{textual_inversion}, Dreambooth~\citep{ruiz2022dreambooth}, and Custom Diffusion~\citep{kumari2023multi} are seminal works in this direction. Recently, Break-a-Scene~\citep{avrahami2023breakascene} introduced masked diffusion loss to learn a new concept, which aligns cross-attention maps with segmentation mask of the given concept. These methods entangle the illuminant from training examples and struggle to illuminate concepts in different illuminants.

\subsection{Illuminant control in image generation.} 
Learning illuminant in image generation is an essential topic~\citep{kocsis2024lightit,butt2024colorpeel}. 
In terms of \textit{color}, most direct option is to apply traditional methods of color constancy~\citep{afifi2020deep, barron2017fast} or color transfer ~\citep{Luan_2017_CVPR} directly to the output, but results are usually not satisfactory. 
Regarding \textit{illuminant}, since T2I diffusion models are proven to be aware of intrinsics~\citep{zeng2024rgbx,kocsis2023intrinsic}, ControlNet~\citep{zhang2023controlnet} method has allowed the appearance of different approaches that take advantage of physical image decomposition for scene relighting~\citep{kocsis2024lightit}. Brook et al.~\citep{brooks2023instructpix2pix} proposed Intruct Pix2Pix which allows to invert images given text prompt. Recently, Xing \emph{et al.} \cite{xing2024retinex} proposed a diffusion model based on Retinex for image relighting. However, little attention has been paid to methods that provide lighting control during text-driven diffusion-based image generation. DiLightNet \cite{zeng2024dilightnet} is pioneer in this topic, but focuses only on the relighting problem under directional lighting. Recently, Zhang et al.~\citep{zhang2025scaling} proposed a diffusion-based image relighting method which allows users to manipulate the original image lighting given text prompt and a light direction. However, model does not preserve spatial background of user-provided image. Here we focus on learning a new concept and synthesizing it into user-guided illumination while preserving all the spatial background information.

\section{Preliminaries}
\label{sec:preliminaries}

\subsection{Latent Diffusion Models.}
In this work, we build on Stable Diffusion adapted through T2I personalization methods for customized T2I generation. It is a Latent Diffusion Model (LDM)~\citep{Rombach_2022_CVPR_stablediffusion} which consists of two main components: (i) An auto-encoder --- that transforms an image $\inputimage$ into a latent code $z_0=\encoder(\inputimage)$ while the decoder reconstructs the latent code back to the original image such that $\decoder(\encoder(\inputimage)) \approx \inputimage$; and (ii) A diffusion model ---
can be conditioned using class labels, segmentation masks, or textual input. Let $\tau_\psi(y)$ represent the conditioning mechanism that converts a condition $y$ into a conditional vector for LDMs. The LDM model is refined using the noise reconstruction loss:
\begin{equation}
    \mathcal{L}_{LDM}\! =\! \expec_{z_0 \sim \encoder(x), \epsilon \sim \mathcal{N}(0, 1)} \left( \Vert \epsilon - \model(z_{t},t, \tau_{\psi}(y)) \Vert_{2}^{2} \right)
    \label{eq:loss}
\end{equation}
The backbone $\model$ is a conditional UNet~\citep{ronneberger2015u} which predicts added noise. Diffusion models aim to generate an image from the random noise $z_T$ and a conditional prompt $\textprompt$. 
We further itemize textual condition as $\textembedding=\tau_\psi(\textprompt)$. 

\subsection{ControlNet.} ControlNet is a neural network extension designed to introduce spatial conditioning into diffusion models, allowing control over image generation~\citep{zhang2023controlnet}. It augments pre-trained diffusion models by incorporating an additional trainable network that conditions outputs based on structured inputs such as edge maps, depth, and more. The loss is given by:
\begin{equation}
    \mathcal{L}_{CN} \!=\! \expec_{z_0 \sim \encoder(x),  \epsilon \sim \mathcal{N}(0, 1)} \underbrace{\left( \Vert \epsilon \text{-} \epsilon_{\theta, \phi}(z_{t},t, \tau_{\psi}(y), C(z_0)) \Vert_{2}^{2} \right)}_{\mathcal{L}_{rec}}
    \label{eq:loss2}
\end{equation}
Here $\phi$ are the additional parameters of the parallel network introduced by the ControlNet, and $C(z_0)$ is the additional structural conditioning. In this paper, we will consider the structural conditioning $C(z_0)$ to be Canny Edge detection.

\subsection{T2I Personalization.}
The task of personalization in T2I diffusion models aims to learn a new concept given a small set of images and text descriptions. \textit{A new concept refers to an entity--such as a person, pet, or object--that user aims to learn and synthesize from text prompts.} Recently, several personalization methods~\citep{textual_inversion,ruiz2022dreambooth,avrahami2023breakascene,kumari2023multi} are introduced that adapt pre-trained diffusion models to learn new concept given a few images along with text descriptions. These methods fine-tune models for new concept while preserving its prior knowledge to ensure generation of newly learned concept given text prompts. As a common practice, these methods introduce a learnable pseudo-word [v] in vocabulary of the model, initialized either by random or specific pre-trained token embeddings. During training, this pseudo-word [v] is optimized using standard diffusion or customized loss functions to learn the embedding of given concept. Though, most of the model parameters remain frozen during training, some methods partially update U-Net to achieve better generation. In this paper, we adapt Custom Diffusion~\citep{kumari2023multi} to learn special tokens for standard lightning conditions, allowing user to precisely control lightning in the generated scene.

\begin{figure}[tb]
  \centering
  \includegraphics[width=0.99\linewidth]{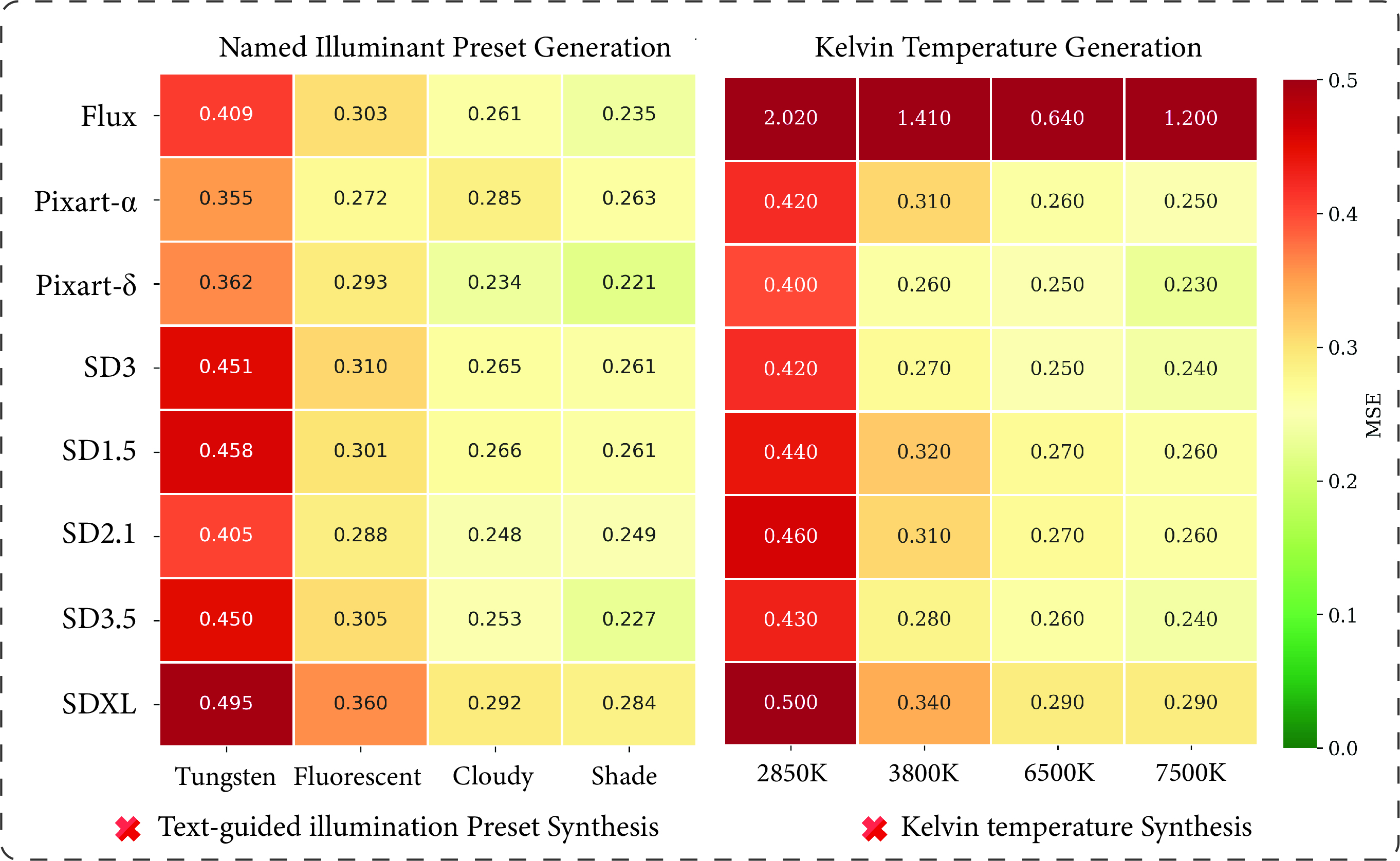}
  \vspace{-2mm}
  \caption{Text-guided illuminant control in T2I models. MSE between estimated and ground-truth illuminated images confirm high error across all models for named presets and kelvin temperatures.}
  \vspace{-6mm}
  \label{fig:heatmap}
\end{figure}

\subsection{Text Encoder Deficit in Illuminant Semantics}
\label{subsec:embedding_analysis}
A fundamental limitation of T2I models is their inability to interpret standard illuminant terms like \emph{tungsten}, \emph{cloudy}, or \emph{6500K} as lighting instructions. To diagnose this, we probe 400 text prompts across 20 objects under four canonical illuminants: \emph{tungsten} (2850K), \emph{fluorescent} (3800K), \emph{cloudy} (6500K), and \emph{shade} (7500K). We generate images with T2I models and apply white-balancing~\citep{afifi2020deep} to estimate implicit scene illuminant. Specifically, we compute per-pixel ratio of original to white-balanced images, aggregate over the foreground mask, and compare the resulting RGB vector to the ground-truth Planckian reference via CIELab MSE.

Figure~\ref{fig:heatmap} reveals a consistent failure that generated images show no systematic shift towards target illuminant. Instead, outputs remain biased toward a default daylight prior, regardless of the prompts. MSE between estimated and ground-truth illuminants is high across all conditions which indicates that model does not respond to illuminant instructions. We observe that this is not due to a lack of visual knowledge, but rather a semantic gap: text encoder does not associate “tungsten” with “warm light,” or its Kelvin equivalent “2850K". To investigate this, we probe text embeddings of four CLIP-based encoders used in T2I pipelines~\citep{cherti2023reproducible} on four token categories: (i) \textit{Standard illuminants} (e.g., \emph{tungsten}), (ii) General lighting terms, (iii) \textit{Kelvin equivalents} (e.g., \emph{2850K}), and (iv)  \textit{Generic numerals} (e.g., \emph{2000}).

\begin{figure}[!t]
  \centering
  \includegraphics[width=0.95\linewidth]{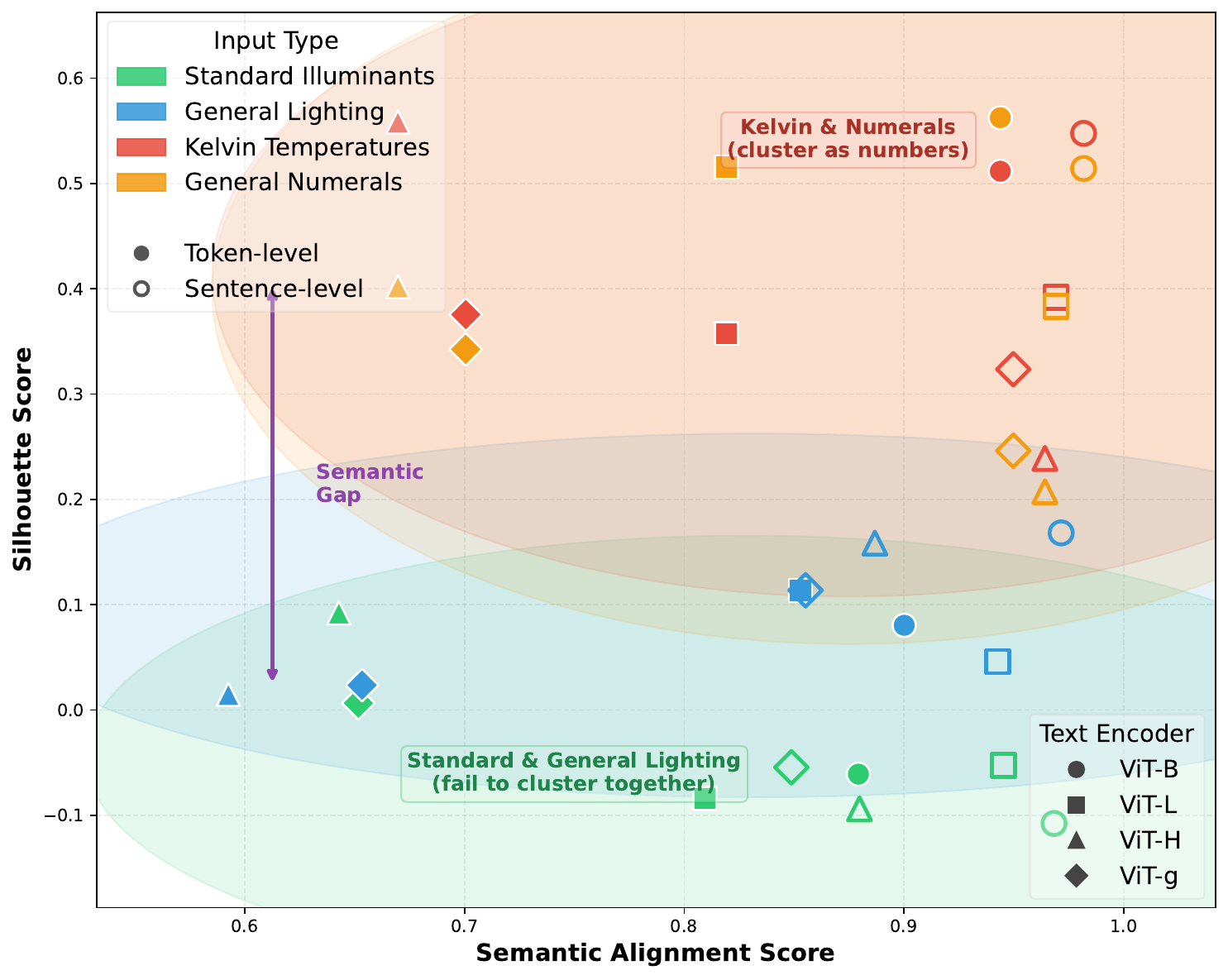}
  \vspace{-2mm}
  \caption{Illuminant embedding analysis across CLIP text encoders. Kelvin values cluster with numerals, confirming they are treated as numbers, not lighting semantics. Standard illuminants and general lighting terms fail to cluster together, revealing semantic gap that explains illuminant prompting failure in T2I models.}
  \label{fig:illum_clustering}
  \vspace{-6mm}
\end{figure}

As shown in Figure~\ref{fig:illum_clustering}, embeddings of standard illuminant presets and kelvin temperatures exhibit two critical failures. First, they do not cluster with general lighting concepts in all models. Second, kelvin temperature values embed near generic numbers rather than with lighting semantics which indicates that encoder treats them as ordinary integers, not photometric quantities. We quantify this semantic disorganization using silhouette scores across four clustering configurations. The scores for “Standard vs. Kelvin” and “Standard vs. General Lighting” are consistently low or negative, confirming that encoders does not recognize semantic equivalence between illuminant names and their physical counterparts or general lighting terms. In contrast, “Kelvin vs. General Numeric” yields high scores, validating that kelvin tokens are interpreted numerically. This analysis reveals a fundamental semantic gap: current text encoders lack the domain-specific grounding required to interpret illuminant instructions. Consequently, T2I methods that rely solely on prompting~\cite{textual_inversion,ruiz2022dreambooth,kumari2023multi} entangle lighting with object identity, as they cannot disentangle illuminant semantics from training data.

\begin{figure*}[t!]
  \centering
  \includegraphics[width=0.90\linewidth]{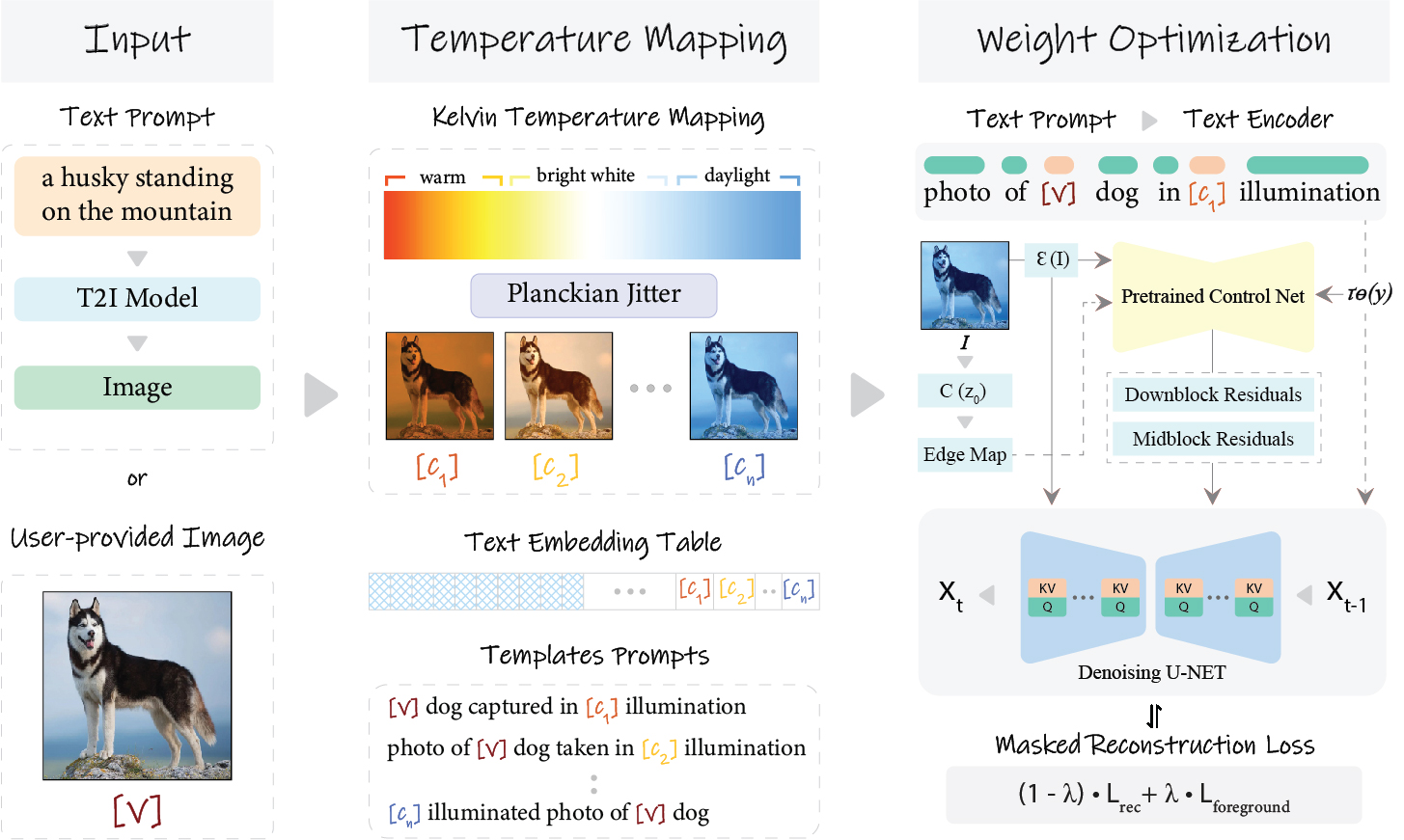}
  \vspace{-2mm}
  \caption{An overview --- \ourmethod consists of three components. Firstly, given an image and text-prompt, our method augments image under daylight illuminants using physics-based color augmentation to learn embeddings. Next, we introduce text-tokens to learn illuminant representations. During training, we only optimize key and value projection matrices in diffusion model cross-attention layers, along with modifier tokens. We employ ControlNet for Edge-Guided Prompt Disentanglement. Third, we introduce masked reconstruction loss to enforce focus in foreground to improve learning. At inference time, ControlNet is discarded.
  }
  \vspace{-6mm}
  \label{fig:method}
\end{figure*}

\section{\ourmethod: Illuminant Prompt Learning}
\label{sec:method}
In this section, we describe \ourmethod to achieve \textit{illuminant prompt learning}, as shown in Figure~\ref{fig:method}. 

\subsection{Temperature mapping.} Computer vision has an extensive literature on illuminant estimation (color constancy)~\citep{gegenfurtner2024color, afifi2019color}. Other than our work, which introduces an illuminant into a scene, these methods study the reverse process of removing illuminant color obtaining the image under white illuminant. Most methods are based on assumption that the scene has a single global illuminant which can be removed by multiply all pixels with a diagonal matrix, where the diagonal is inverse of the illuminant color. The reverse process, which we call \emph{flat light adaptation}, can hence be obtained by multiplying the pixels with a diagonal matrix with diagonal equal to the desired scene illuminant. It was found that realistic illuminants can be modeled by Planckian locus (or black body locus)~\citep{zini2022planckian}. 
here, we consider $N=7$ illuminants. Four of them are traditional presets found in cameras and post-capture softwares: $c_1=$\quotesingle{Tungsten}(2850K), $c_3=$\quotesingle{Fluorescent}(3800K), $c_5=$\quotesingle{Cloudy}(6500K) and $c_7=$\quotesingle{Shade}(7500K). The other three are the intermediate illuminants: $c_2=$ 3300K, $c_4=$ 4500K, and $c_6=$ 7000K. The flat light adaptation applies same illuminant transformation to the whole image. Here, we use flat light adaptation to provide training data for illuminant prompts in temperature mapping step. We will introduce an additional loss that focuses on color transformation of foreground object, and we will leave background illuminant generation to the prior knowledge inherent to the pretrained diffusion model.

\subsection{Weight optimization.} Following the principles of T2I personalization, we introduce text tokens $\concepttoken$ corresponding to the embeddings of the original concept. To learn the illuminant representation, we introduce separate tokens --- $\illumtoken$ for each illuminant. To further improve learning, we create a small subset of text prompts such as \quotes{a photo of $\concepttoken$ dog in $\illumtoken$ illuminant} to condition the model during training. To learn these illuminant tokens concept $\concepttoken, \illumtoken$, an image is sampled along with the text prompt from the given set of images and their prompt templates, and $\concepttoken, \illumtoken$ are directly optimized by mitigating the loss of LDM as defined in Eq.~\ref{eq:loss}. Therefore, our optimization function can be defined as:

\begin{equation}
\{\concepttoken, \illumtoken \} = \underset{\tokenset}{\arg\min} \   \expec_{z_0 \sim \encoder(\inputimage_i), y, \epsilon \sim \mathcal{N}(0, 1)} \; \mathcal{L}_{rec} 
\label{eq:token_loss}
\end{equation}

where the similar training scheme of the actual LDM is reused to optimize $\concepttoken$, allowing learned embedding to capture fine visual details of the given illuminant concept.

\subsection{Edge-Guided Prompt Disentanglement.} A major challenge of illuminant prompt learning is \emph{spurious detail leakage}, where prompt inadvertently captures structural elements of training images. Consequently, the image content can undergo structural changes including the introduction or removal of objects during generation. 
To address this, we introduce a pre-trained ControlNet conditioned on edge maps with frozen parameters, providing edge guidance so that illuminant prompt learning focuses solely on color information, as the edge maps already encode structural information.
We note that this effectively addresses the spurious detail leakage problem, and learned illuminant prompts no longer alter image structure. Note that the ControlNet is not applied at inference time.

\begin{figure*}[t!]
  \centering
  \includegraphics[width=0.95\linewidth]{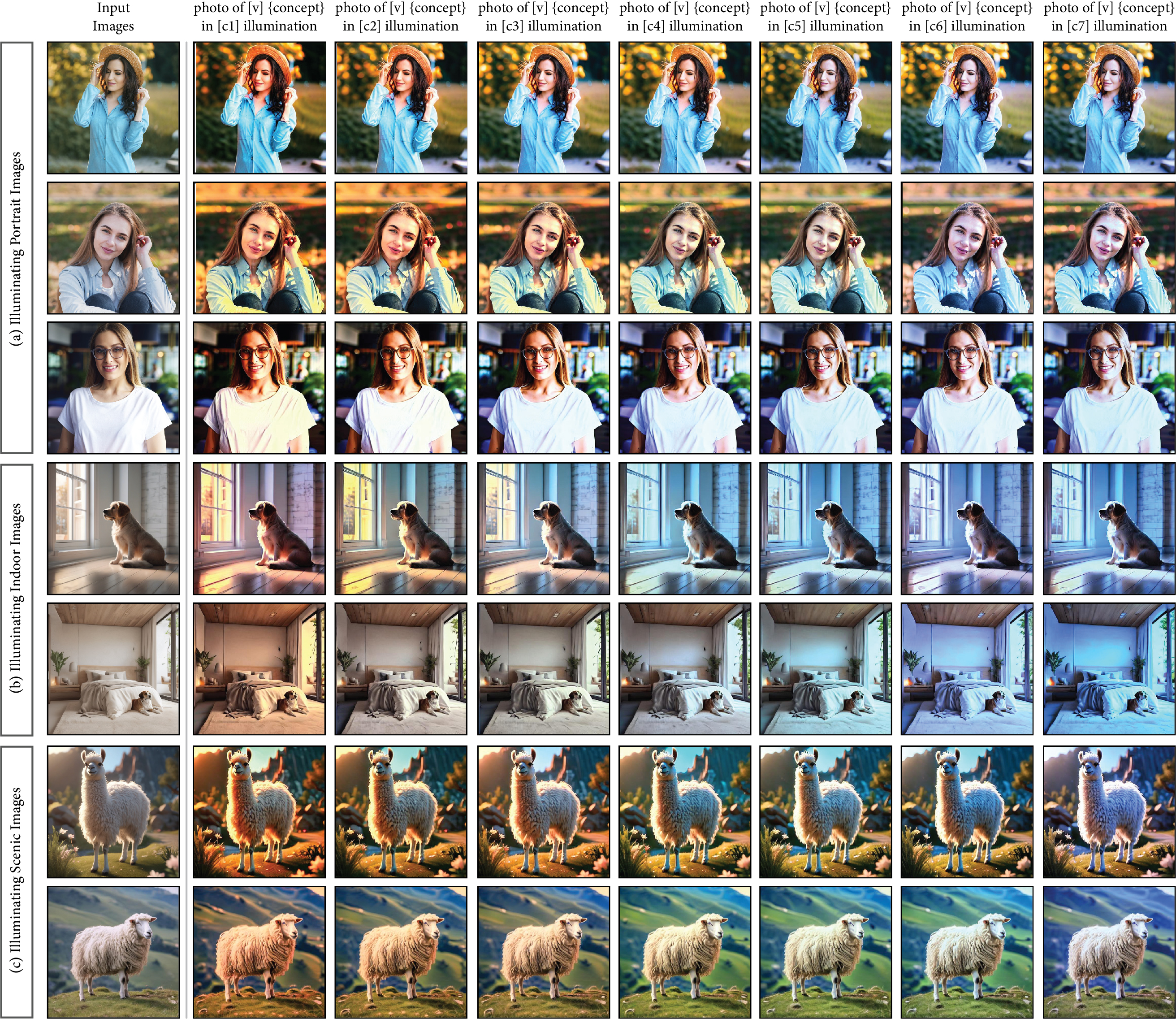}
  \caption{Qualitative results of \ourmethod illuminating concepts under three settings: (a) Portrait, (b) Indoor, and (c) Outdoor illumination.
  }
  \vspace{-3mm}
  \label{fig:qualitative_results}
\end{figure*}

\subsection{Masked reconstruction loss.}
\label{subsubsec:illum_disent}
Flat Light Adaptation applies uniform transformation on the image to manipulate illuminant, which makes an image unrealistic, whereas real-world illuminant changes often affect different parts of an image unevenly due to shadows, reflections, and varying material properties. Consequently, this method~\citep{zini2022planckian} hinders in correctly transforming an image while considering the relation of light and surfaces, potentially limiting the generalization of concepts in real-world scenarios. To address this, we propose \textit{Masked Reconstruction Loss} (MRL) to penalize more over the pixels covered by the mask of the concept while less focus on background of the image. MRL is adapted as:
\begin{align}
\mathcal{L}_{mrl} = \underbrace{(1-\lambda) \cdot \mathcal{L}_{rec} \cdot (1-\mask)}_{background} + \underbrace{\lambda \cdot \mathcal{L}_{rec} \cdot \mask}_{forground}
\label{eq:mrl_loss}
\end{align}
where $\lambda$ is a trade-off hyperparameter to balance weight between foreground concept and background in the image. By this means, our \ourmethod learns the illuminant prompt by loss $\mathcal{L}_{mrl}$ and precisely captures the illuminant attributes.

\section{Experiments}
\label{sec:experiments}
\subsection{Experiment Setup}
\label{subsec:exp_setup}
\noindent\textbf{Implementation Details.} Following the practices adopted in T2I personalization methods, we conducted all experiments using Stable Diffusion v1.5~\citep{Rombach_2022_CVPR_stablediffusion} as a pre-trained diffusion model on NVIDIA A40 GPU. We curated a set of 20 concepts to learn illuminant prompts from \textit{Real}, and \textit{Generated Images}. \ourmethod is trained using AdamW optimizer with batch size of 2 and learning rate of $10^{-5}$ for 3000 steps. Masked reconstructed loss is computed on 64$\times$64 resolution---the size of image latents. For inference, we set DDPM steps to 200 and classifier-free guidance scale to 6.0.

\noindent\textbf{Comparison methods.}
We compare \ourmethod with state-of-the-art methods for prompt-based illumination control in two categories: (1) T2I personalization and (2) T2I appearance editing. For personalization methods, we evaluate Textual Inversion~\citep{textual_inversion}, DreamBooth~\citep{ruiz2022dreambooth}, Custom Diffusion~\citep{kumari2023multi}, and Break-a-Scene~\citep{avrahami2023breakascene}. For appearance editing methods, we include IC-Light~\citep{zhang2025scaling}, Instruct Pix2Pix~\citep{brooks2023instructpix2pix}, RGB2X~\citep{zeng2024rgbx}, and three Prompt-to-Prompt variants: DDIM+P2P~\citep{hertz2022prompt}, Null-Text+P2P~\citep{avrahami2022blended}, and PnP+P2P~\citep{ju2023direct}. All methods are evaluated under identical conditions using the same concepts and illuminant prompts.

\noindent\textbf{Evaluation Metrics.} Evaluating illuminants in T2I generation is challenging due to lighting variations and reflections. We adopt a color-constancy approach: for each generated image, we apply a white balance method~\citep{afifi2020deep} to produce a neutral illuminant version, then compute the per-pixel ratio between original and white-balanced images, aggregated over a foreground mask. This estimated illuminant vector is compared against the ground-truth Planckian reference using: (i) Mean Angular Error (MAE) for chromaticity deviation; (ii) Mean Squared Error (MSE) for vector accuracy; and (iii) SSIM for structural preservation. All metrics are evaluated on the foreground region. We generated images with 42 random seeds per comparison and conducted a human study between \ourmethod and baselines.

\subsection{Qualitative Analysis}
We evaluate the qualitative performance of \ourmethod and baseline methods in illuminating concepts in both real and T2I generated images, given a set of text prompts under seven different illuminations, ranging from \emph{tungsten} to \emph{shade}. The results in Figure~\ref{fig:qualitative_results} show that \ourmethod is able to capture visual details of target concept from both the real and T2I generated images, and generated concepts with better illumination which aligns well with text prompts.

\begin{figure*}[t!]
  \centering
  \includegraphics[width=0.90\linewidth]{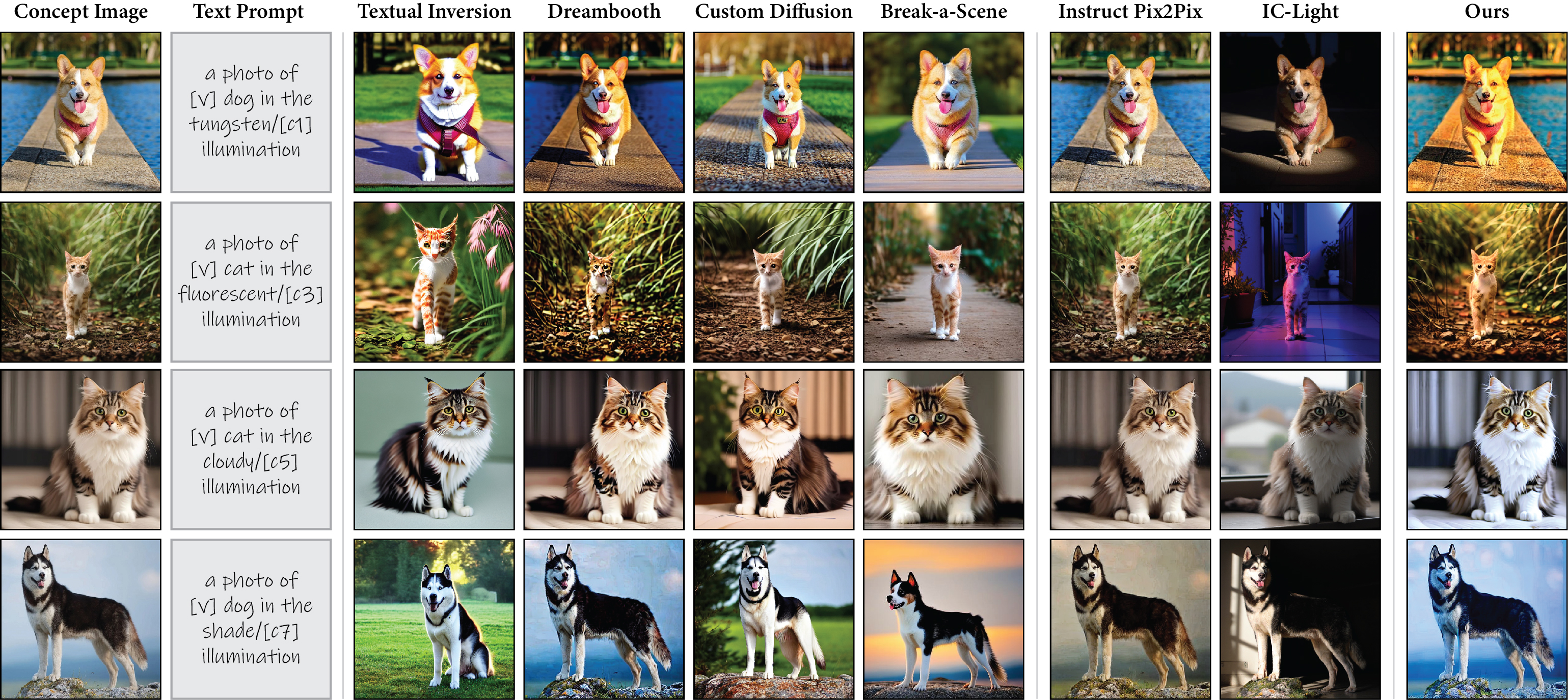}
  \vspace{-2mm}
  \caption{Qualitative results on \textit{illuminant prompt learning} task compared with the baseline T2I personalization methods.
  {Though the baseline methods preserve target concepts, they struggle to synthesize in the given illumination as guided in the text prompt. Whereas, \ourmethod can efficiently synthesize the target concept under different illuminations.}}
  \label{fig:qualitative_comparison}
  \vspace{-6mm}
\end{figure*}

Next, we qualitatively compare \ourmethod and the baseline methods---including T2I personalization methods and image editing methods---on diverse concepts under four standard illuminants. The results in Figure~\ref{fig:qualitative_comparison} demonstrates that \ourmethod successfully generates images where target concept is rendered under the specified illuminant while preserving its identity, texture, and pose. More importantly, \ourmethod achieves contextual light adaptation: the background lighting shifts naturally to complement foreground illuminant, resulting in coherent and photo-realistic scenes. 


In contrast, T2I personalization methods preserve the concept's identity but fail to adapt to text-guided illuminants, consistently retaining lighting conditions from training examples regardless of the prompt. This confirms our hypothesis that these methods entangle object identity with scene illumination. Image editing methods such as Instruct Pix2Pix~\citep{brooks2023instructpix2pix} and IC-Light~\citep{zhang2025scaling} yield unsatisfactory results. Instruct Pix2Pix frequently distorts or introduces visual artifacts. IC-Light demonstrates better performance in controlling light direction, however, it fails to preserve spatial background when foreground conditioned, and introduces inconsistent lighting when conditioned with background. Notably, \ourmethod avoids detail leakage through edge-guided prompt disentanglement and prevents uniform color shifts via masked reconstruction loss, producing contextually grounded images. Ablation studies and quantitative metrics confirm that \ourmethod outperforms baselines in both illuminant accuracy and image fidelity.

\subsection{Quantitative Analysis}
We evaluate \ourmethod on a set of 20 concepts---portraits of humans and pets across various indoor and outdoor settings under seven distinct illuminants. For each method, we generate images using 42 random seeds and compute four quantitative metrics (see Sec.~\ref{subsec:exp_setup}) to assess illuminant accuracy and image fidelity. The results in Table~\ref{tab:quan_results} demonstrate that \ourmethod outperforms baselines across all metrices. Importantly, \ourmethod achieves lowest angular error and MSE, indicating the precision to render chromaticity and intensity of the target illuminant. This confirms that \ourmethod learns precise, grounded illuminant representations, unlike baseline methods whose estimates remain far from the ground-truth. In terms of image quality, \ourmethod also achieves highest SSIM among all methods. This reflects its ability to preserve structural details and perceptual coherence while adapting lighting. 
Notably, our ablation reveals critical role of edge-guidance: \ourmethod without ControlNet still improves over baselines but suffers from higher MSE, confirming that edge guidance is essential for maintaining structural integrity during illuminant editing.

\begin{figure}[!t]
  \centering
  \includegraphics[width=0.9999\linewidth]{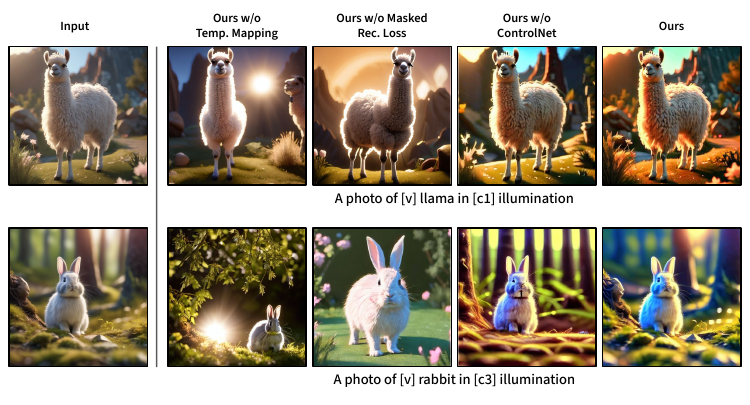}
  \vspace{-8mm}
  \caption{Ablation Study. (i) Removing Temperature Mapping and Masked Reconstruction Loss introduces divergent lighting, causing misalignment with text prompts. (ii) Removing ControlNet guidance introduces artifacts in generated images.}
  \label{fig:ablation_study}
  \vspace{-5mm}
\end{figure}

\begin{table}[t!]
\centering
\scriptsize
\caption{Quantitative comparison of illumination-preserving image editing methods. 
Angular Error and MSE measure the accuracy of estimated illuminant, while 
SSIM measures image fidelity. 
Color legend: 
\colorbox{rank1}{\rule{0pt}{6pt}\rule{10pt}{0pt}} best, 
\colorbox{rank2}{\rule{0pt}{6pt}\rule{10pt}{0pt}} second, 
\colorbox{rank3}{\rule{0pt}{6pt}\rule{10pt}{0pt}} third.}
\label{tab:quan_results}
\resizebox{\linewidth}{!}{%
\begin{tabular}{llccc}
\hline
Category & Method & Angular Error $\downarrow$ & SSIM $\uparrow$ & MSE $\downarrow$ \\
\hline
\multirow{4}{*}{T2I Personalization} 
& Textual Inversion & 15.35 & 0.57 & 38.50 \\
& DreamBooth & 12.76 & \cellcolor{rank3}0.71 & 34.10 \\
& Custom Diffusion & 13.34 & 0.61 & 38.20 \\
& Break-a-Scene & 14.57 & 0.63 & 33.80 \\
\hline
\multirow{6}{*}{T2I Appearance Editing} 
& IC-Light & \cellcolor{rank3}10.39 & 0.58 & 35.90 \\
& Instruct Pix2Pix & 13.67 & 0.60 & 37.00 \\
& RGB2X & 20.68 & 0.61 & 41.20 \\
& DDIM+P2P & 15.03 & 0.68 & 38.40 \\
& Null-Text+P2P & 12.91 & 0.70 & 33.50 \\
& PnP+P2P & 11.24 & 0.67 & \cellcolor{rank3}33.20 \\
\hline
\multirow{2}{*}{Proposed} 
& Ours w/o CtrlNet & \cellcolor{rank2}\underline{6.87} & \cellcolor{rank2}\underline{0.74} & \cellcolor{rank2}\underline{22.40} \\
& Ours w/ CtrlNet & \cellcolor{rank1}\textbf{4.51} & \cellcolor{rank1}\textbf{0.77} & \cellcolor{rank1}\textbf{16.80} \\
\hline
\end{tabular}
}
\vspace{-5mm}
\end{table}


\noindent\textbf{Ablation Study.}
We conduct ablation study over various factors, as shown in Fig.~\ref{fig:ablation_study}. We remove temperature mapping and masked reconstruction loss and note that \ourmethod though preserves target concept; however, it introduces divergent lighting sources when temperature mapping and masked reconstruction loss are removed. Resultantly, images do not align with text-guided illumination. Moreover, \ourmethod introduces artifacts when ControlNet based guidance is removed during training. \ourmethod generates unrealistic lighting and also affects background when $\lambda$ is scaled higher in masked reconstructed loss (see supplements).





\begin{figure}[!t]
  \centering
  \includegraphics[width=\linewidth]{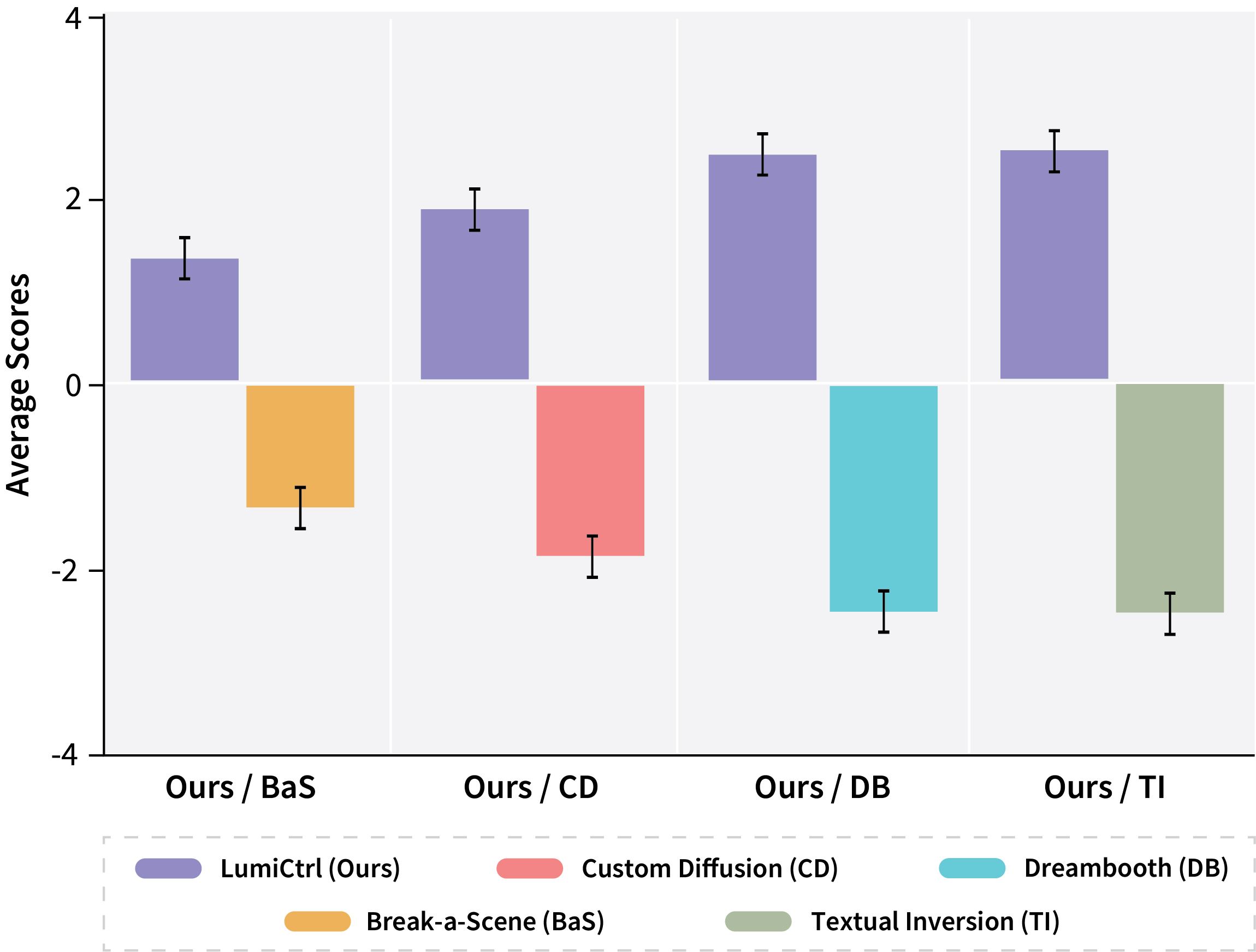}
  \vspace{-5mm}
  \caption{Human preference study using 2AFC protocol with Thurstone's Case V model. Higher z-scores indicate better.}
  \label{fig:human_study}
  \vspace{-4mm}
\end{figure}

\noindent\textbf{User Study.}
We conducted user study with 15 participants to perceptually evaluate our results, comparing \ourmethod with baselines. The experiment was carried out in a controlled environment using two-alternative forced choice (2AFC) method where observers were presented with three images. All observers were tested for correct color vision using Ishihara test. We tested $20$ different concepts and $4$ standard illuminants, accumulating to 320 questions. We compare \ourmethod with T2I personalization methods using Thurstone Case V Law of Comparative Judgment model \cite{thurstone2017}. This method provided us with z-scores and a $95\%$ confidence interval, calculated using method proposed in \cite{montag2006}. The results are shown in Fig.~\ref{fig:human_study}. \ourmethod outperforms competing methods which underscores the effectiveness of \ourmethod in generating realistic illuminant images.

\minisection{Limitations.}
While \ourmethod achieves high-fidelity with \textit{illuminant prompt learning}, it is not free of limitations. We focus on seven widely adopted illuminants to establish a reproducible foundation, however, content creators could require more rare illuminants which we have not been taken into account. Additionally, as illumination exists on a continuous spectrum, discrete token additions to the CLIP vocabulary provide a baseline but can be further explored. Developing a unified training framework capable of capturing this continuous variability can be an interesting direction.

\section{Conclusion}
\label{sec:conclusion}
We identified a semantic gap in T2I models that standard illuminants are not grounded in text encoder embeddings, causing models to default to daylight. We proposed \ourmethod, first prompt-space illumination method that learns \emph{illuminant prompts} from a single concept image. By integrating physics-based augmentation, Edge-Guided Prompt Disentanglement, and Masked Reconstruction Loss, \ourmethod achieves precise, context-aware lighting control while preserving object identity and spatial structure. Experiments and human study confirm \ourmethod outperforms existing methods in text-guided illuminant generation.

\section*{Acknowledgments}
This work was supported by Grants PID2021-128178OB-I00, PID2022-143257NB-I00, AIA2025-163919-C52, and PID2024-162555OB-I00 funded by MICIU/AEI/10.13039/501100011033, ERDF/EU and the FEDER, by the Generalitat de Catalunya CERCA Program, by the grant Càtedra ENIA UAB-Cruïlla (TSI-100929-2023-
2) from the Ministry of Economic Affairs and Digital Transition of Spain, by the European Union’s Horizon Europe research and innovation programme under grant agreement number 101214398 (ELLIOT), and by the BBVA Foundations of Science program on Mathematics, Statistics, Computational Sciences and Artificial Intelligence (grant VIS4NN). JVC also acknowledges the 2025 Leonardo Grant for Scientific Research and Cultural Creation from the BBVA Foundation. The BBVA Foundation accepts no responsibility for the opinions, statements and contents included in the project and/or the results thereof, which are entirely the responsibility of the authors. Kai Wang acknowledges the funding from Guangdong and Hong Kong Universities 1+1+1 Joint Research Collaboration Scheme and the start-up grant B01040000108 from CityU-DG.

{
    \small
    \bibliographystyle{ieeenat_fullname}
    \bibliography{main}
}

\clearpage
\appendix

\twocolumn[
\centering
{\Large \bfseries Supplementary Material \par}
\vspace{0.5em}
\vspace{1.5em}
]

\vspace{2em}

\renewcommand{\thesection}{S\arabic{section}}
\setcounter{section}{0}

\renewcommand{\thesubsection}{S\arabic{section}.\arabic{subsection}}

\renewcommand{\thefigure}{S\arabic{figure}}
\setcounter{figure}{0}

\renewcommand{\thetable}{S\arabic{table}}
\setcounter{table}{0}

\section{Learning Illuminants with T2I Personalization Methods}
Text-to-image (T2I) diffusion models allow users to synthesize objects by incorporating linguistic descriptions into text prompts, such as \quotes{a cat sitting on the mountain} or \quotes{a fashion model walking on the ramp}. Although these T2I diffusion models have demonstrated remarkable abilities in generating images from text prompts, they struggle with synthesizing objects in precisely user-guided illuminations.

Recently, T2I personalization methods have been proposed, including Textual Inversion~\cite{textual_inversion}, Dreambooth~\cite{ruiz2022dreambooth}, Custom Diffusion~\cite{kumari2023multi}, and Break-a-Scene~\cite{avrahami2023breakascene}. These personalization methods allow users to learn personal concepts such as friends, pets, or specific items given 4-5 images. We employ these methods to learn concepts such as cats and dogs and evaluate the performance in terms of illuminating the newly learned concept under different illuminations using text prompts: a photo of [v] dog under tungsten/fluorescent illumination. The results demonstrated in our main paper show that although these methods efficiently preserve the identity of the given concept, they struggle to illuminate concepts under text-guided illumination.

\begin{figure}[!t]
  \centering
  \includegraphics[width=0.95\linewidth]{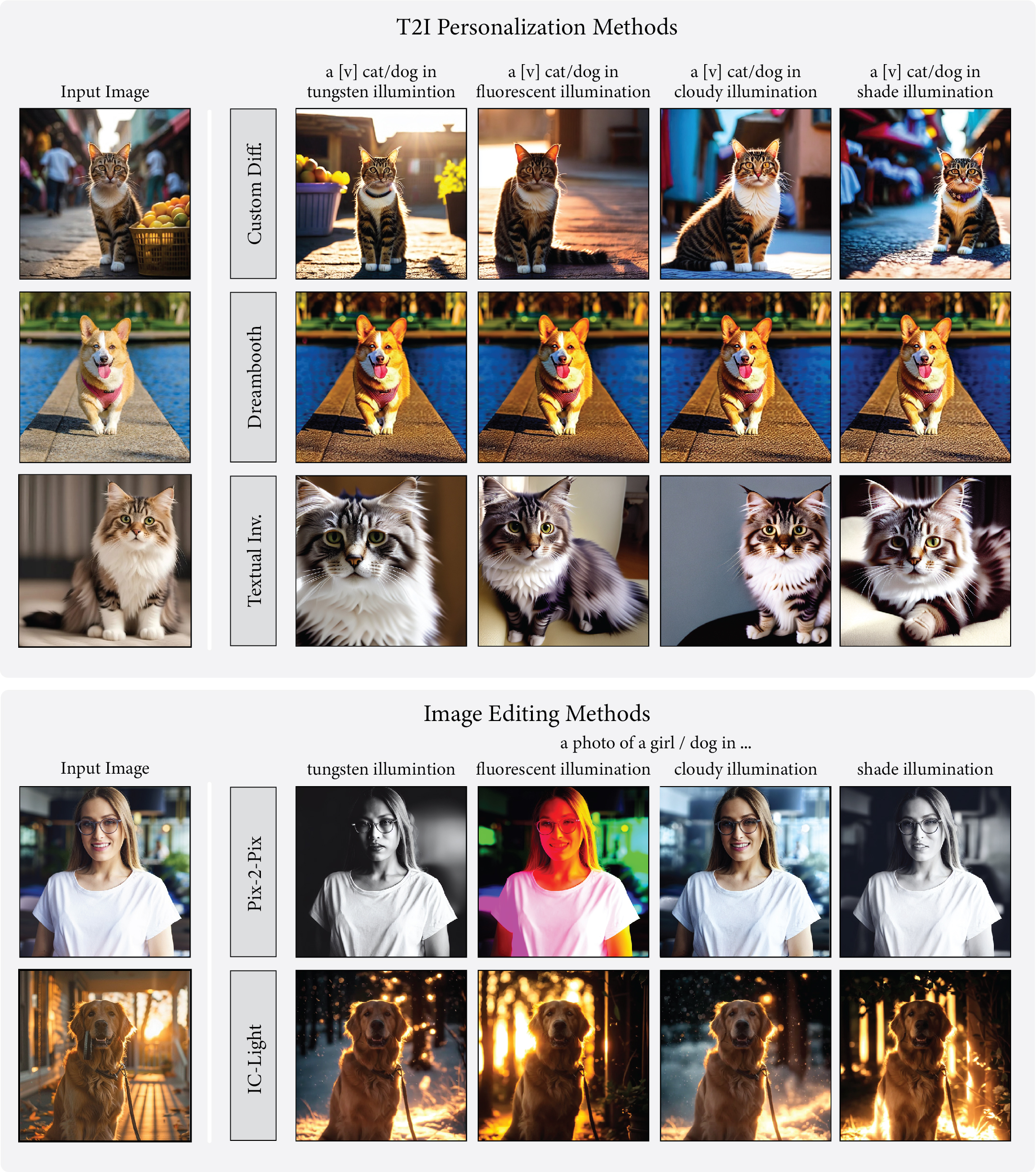}
  \caption{Analyzing the capability of T2I personalization methods, and Image Editing Methods in illuminating concepts using text-prompts. \underline{\textbf{T2I Personalization Methods}}: Textual Inversion, Dreambooth, and Custom Diffusion predominantly preserve lighting from training examples, failing to generate concepts under various illuminants. \underline{\textbf{Image Editing Methods}}: These methods fails to photo-realistically change the illumination aligning with the text prompt. Note that, the IC-Light generations are from foreground conditioning.}
  \label{fig:baseline_comp_supp}
\end{figure}

In particular, Custom Diffusion, Dreambooth, and Break-a-scene tend to synthesize concepts in a similar posture and adopt daylight illumination given in the example images. This behavior stems from the entanglement of object identity with scene illumination during fine-tuning—the model learns to associate the concept with its original lighting conditions, making it resistant to illuminant changes at inference time. Textual Inversion, on the other hand, does not synthesize the concept faithfully and fails to adopt the texture from training examples. Resultantly, these methods do not synthesize concepts under text-guided illumination, as evident in Fig.~\ref{fig:baseline_comp_supp}.

\begin{figure}[!t]
\centering
\includegraphics[width=0.999\linewidth]{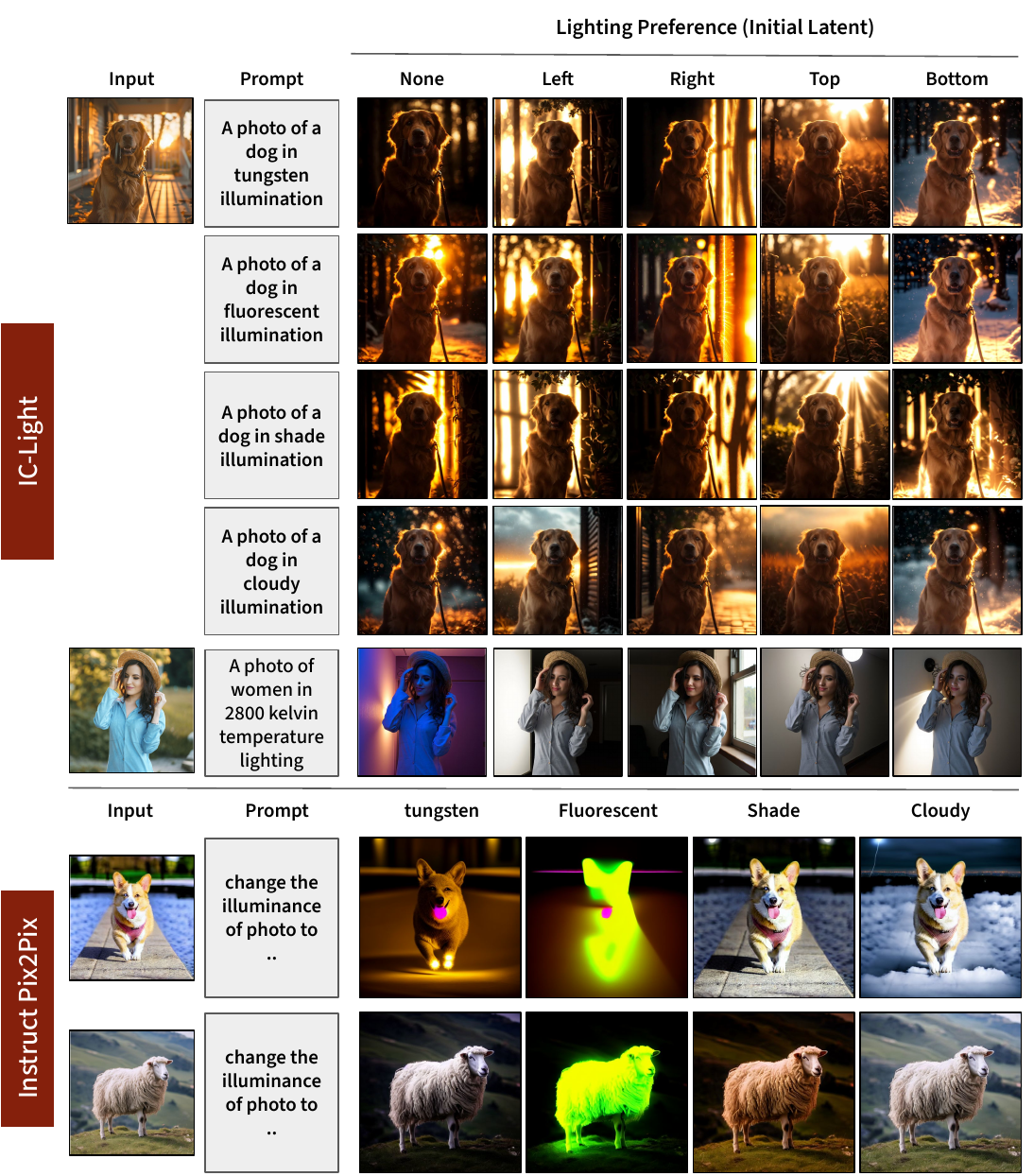}
\caption{Inferences with IC-Light and Instruct Pix2Pix. IC-Light struggles with preserving the spatial background information. Moreover, both models fail to understand the text-guided daylight illuminants and kelvin temperature. IC-Light examples are from foreground conditioning.}
\label{fig:ic_light}
\end{figure}

We also analyze the performance of image editing methods, i.e., Pix2Pix~\cite{brooks2023instructpix2pix} and IC-Light~\cite{zhang2025scaling}, showing results in Fig.~\ref{fig:baseline_comp_supp} and Fig.~\ref{fig:ic_light}. We employ pre-trained baselines and generate images given input image and text prompt. It can be observed that Pix2Pix and IC-Light struggle to synthesize concepts under text-guided illumination. Though IC-Light demonstrates considerably better performance in manipulating lighting direction, it has two key limitations. First, it does not preserve the spatial background information of the given image, often introducing inconsistent geometry or artifacts. Second, it struggles to understand and adapt to text-guided daylight illuminants and Kelvin temperatures, as these terms lack semantic grounding in the underlying text encoder—a fundamental limitation we identify and address with LumiCtrl.

\section{Experiments}

\subsection{Training Examples}
\begin{figure}[!t]
\centering
\includegraphics[width=0.979\linewidth]{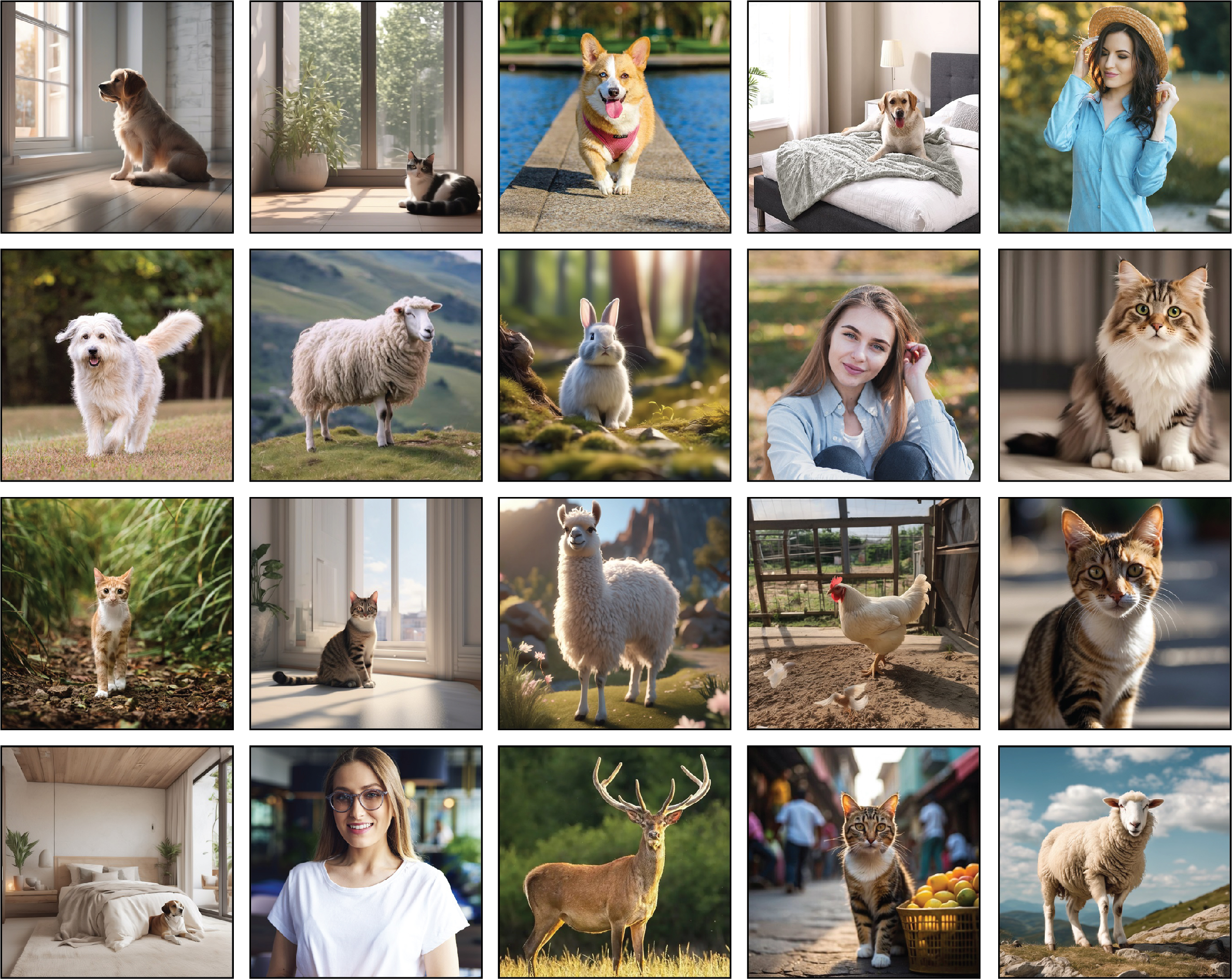}
\caption{Data Samples --- Concepts used in the qualitative and quantitative study.}
\label{fig:dataset}
\end{figure}

In this work, we curated a small set of 20 concepts to learn illuminant prompts from \textit{Real Images}, and \textit{Generated Images}, shown in Fig.~\ref{fig:dataset}. For real images, we pick images of pets including cats and dogs from multiple internet sources~\cite{chien_2024,Photography_2021,KHARB, freepik}, and we used different text prompts to generate training concepts, including llama, rabbit, and cows.
We provide a list of text prompts used to generate these images in Section~\ref{subsec:train_prompt}.

\subsection{Training Prompts}
\label{subsec:train_prompt}

We generate training examples with text prompts as below.

\begin{itemize}
    \item a realistic photo of a walking Siberian Husky in the grass field.
    \item a photo of a cute realistic llama, highly detailed, cinematic illuminant.
    \item a photo of a white cow walking through the grass field.
    \item a photo of a heartwarming adorable rabbit sitting in the meadows.
    \item a photo of a white horse standing in front of a house.
    \item a photo of a chicken playing in the garden.
    \item a photo of a white sheep on the mountain.
    \item a photo of a rabbit in the garden.
    \item a photo of a cat sitting by the window in a room.
    \item a photo of a dog sitting on the floor by the window in an ultra realistic modern room.
\end{itemize}

\subsection{Evaluation Settings}
We optimized the new text tokens with multiple training prompt examples, listed below.

\begin{itemize}
      \item a photo of $[v]$ {concept} captured in [$c_*$] illumination.
      \item a photo of $[v]$ {concept} in [$c_*$] illumination.
      \item $[v]$ {concept} in [$c_*$] illumination.
      \item a photo of $[v]$ {concept} taken in [$c_*$] illumination.
      \item a [$c_*$] illuminated photo of $[v]$ {concept}.
      \item a photo of $[v]$ {concept} with [bg] background captured in [$c_*$] illumination.
      \item a photo of $[v]$ {concept} with [bg] background taken in [$c_*$] illumination.
      \item a [$c_*$] illuminated photo of a $[v]$ {concept} with [bg] background.
\end{itemize}

 \noindent Here we use the aforementioned set of instance prompts, where we learn embedding of the concept such as dog or cat in $[v]$ text-token, [bg] as background, and [c1],[c2],[c3],[c4],[c5],[c6], and [c7] text-tokens to learn illumination embeddings of tungsten, fluorescent, cloudy, and shade, and the three intermediate illuminants respectively.

\subsection{User Study}
\begin{figure}
\centering
\includegraphics[width=\linewidth]{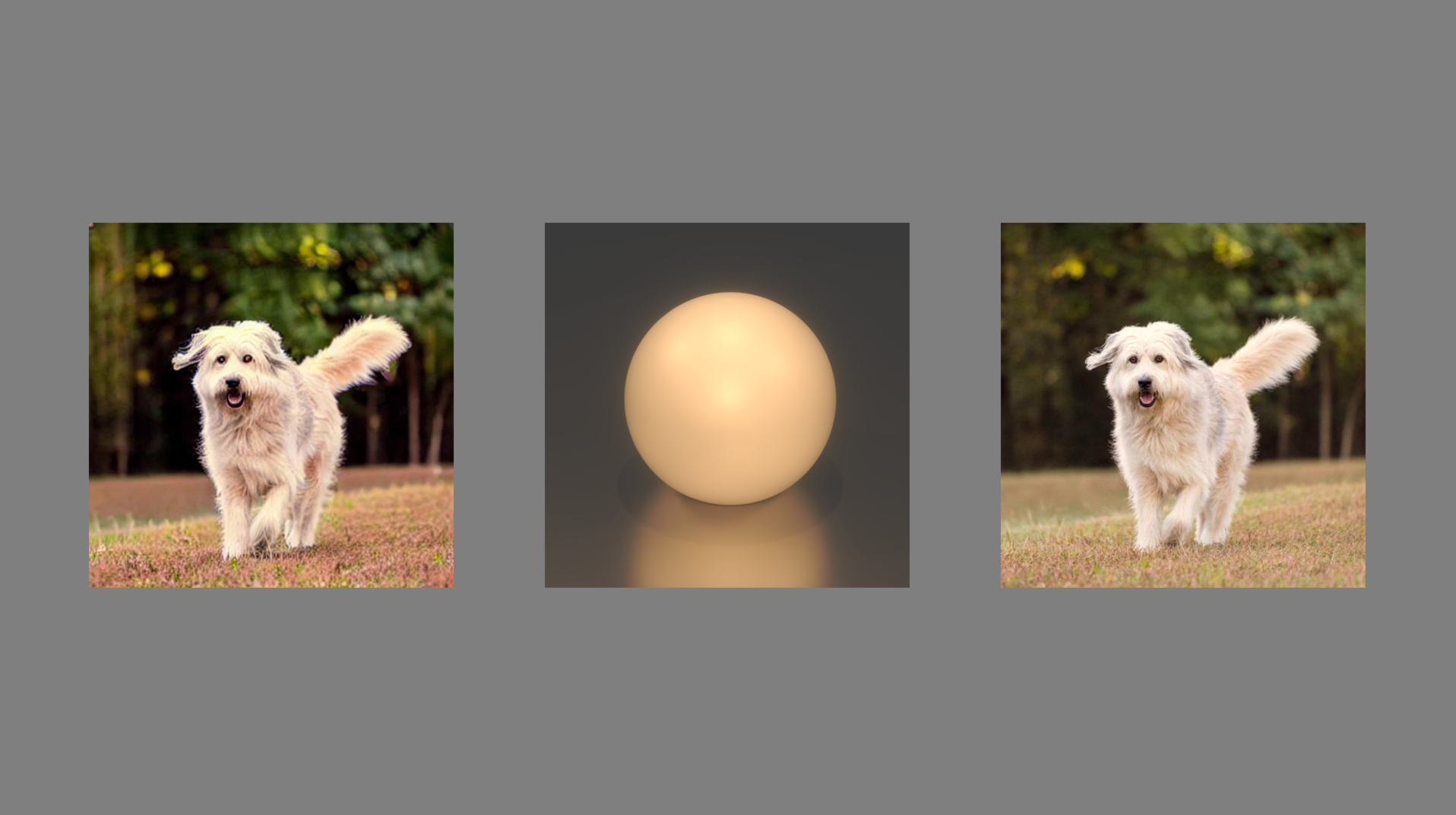}
\caption{Setup of our User Preference Study: The monitor's background was set to a neutral gray, and participants were asked to choose which of the two images — left or right — best matched the illuminant color, based on the color displayed in the sphere in the central image.}
\label{fig:human_study_setup}
\end{figure}

 The room in our user study was completely dark, with the monitor set to sRGB mode. The only light source during the experiment was the monitor. Participants were advised to sit about 60 cm away from the monitor, providing a 7-degree visual angle. The monitor background was set to the neutral gray, displaying a central image containing an sphere representing the light color. On either side of this image, we randomly presented results from our method and a competing one. The participant’s task was to choose which of the two images best matched the prompt based on the illuminant color in the central image. Fig.\ref{fig:human_study_setup} shows an example of what the observer saw. A total of 15 participants participated in the study, none of the authors involved. The central image presents a gray sphere illuminated by the color of illuminant name ---i.e. how a sphere will look under that illuminant name. To left and right, we show results of our method and one of the competing methods, in randomized order.

\subsection{Comparisons versus Flat Light Algorithms}
\begin{figure}
\centering
\includegraphics[width=\linewidth]{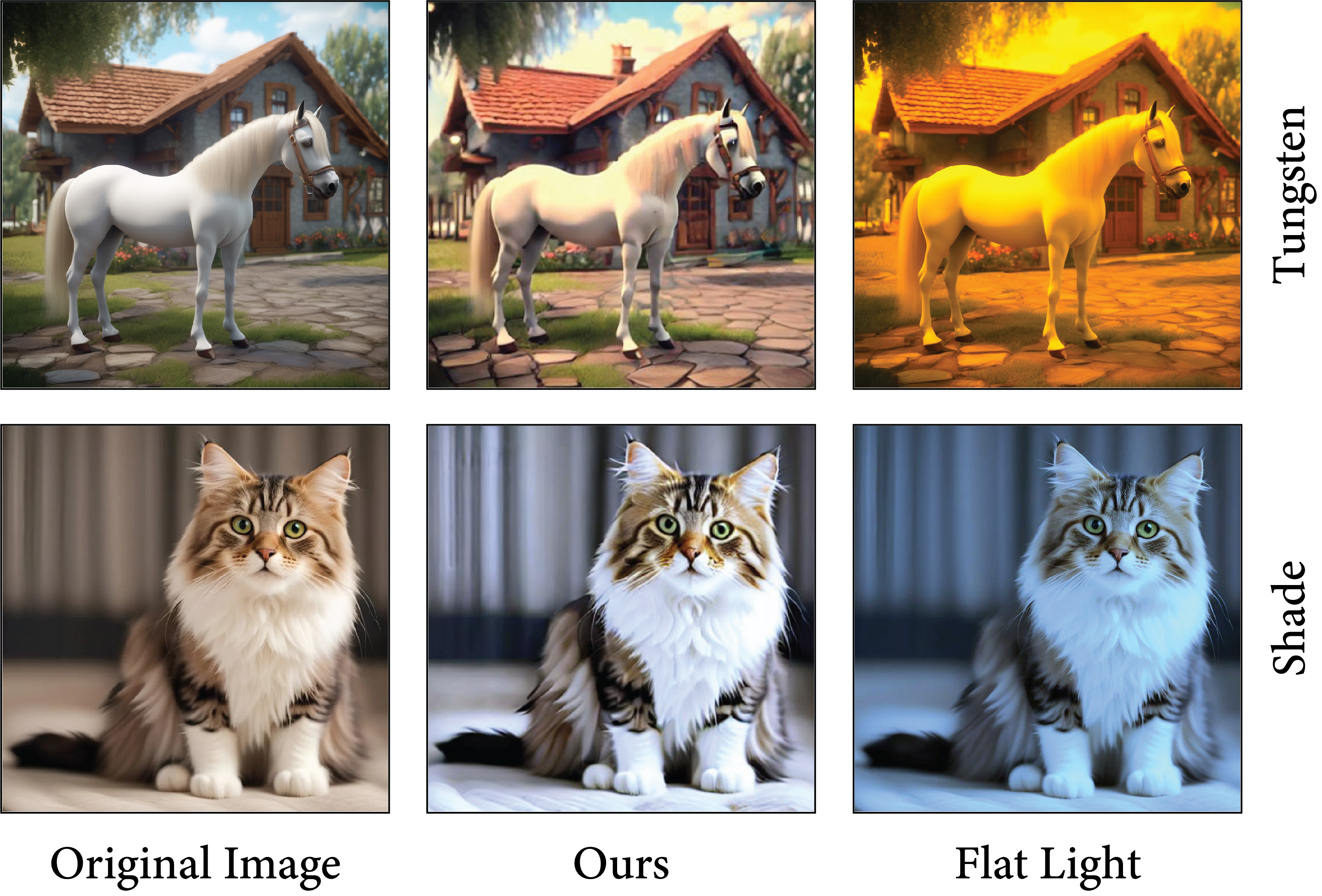}
\caption{Comparison between our approach and the Flat Light Adaptation. Flat Light Adaptation leads to unrealistic results, while our method provides pleasant images.}
\label{fig:illum_flat_ours}
\end{figure}

\begin{figure}[!t]
\centering
\includegraphics[width=0.99\linewidth]{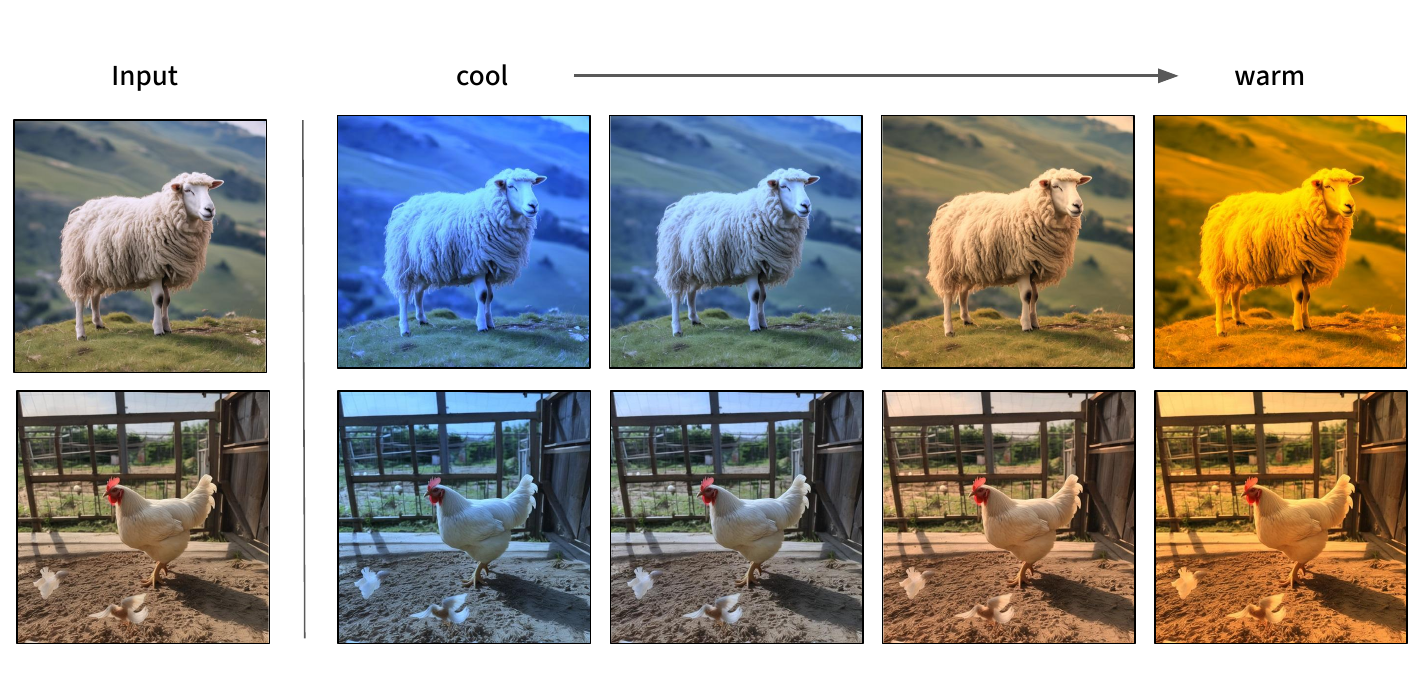}

\caption{Demonstration of illuminating concepts into different illuminant conditions using Flat Light Adaptation. It can be noted that image becomes more unrealistic, when the illuminant is scaled higher in both--- the cool, and warm conditions. Secondly, Flat Light Adaptation cannot create illuminance adaptive shadow, which is one of the main drawbacks of this approach.}

\label{fig:illum_fla}
\end{figure}

\begin{figure*}[!t]
\centering
\includegraphics[width=\linewidth]{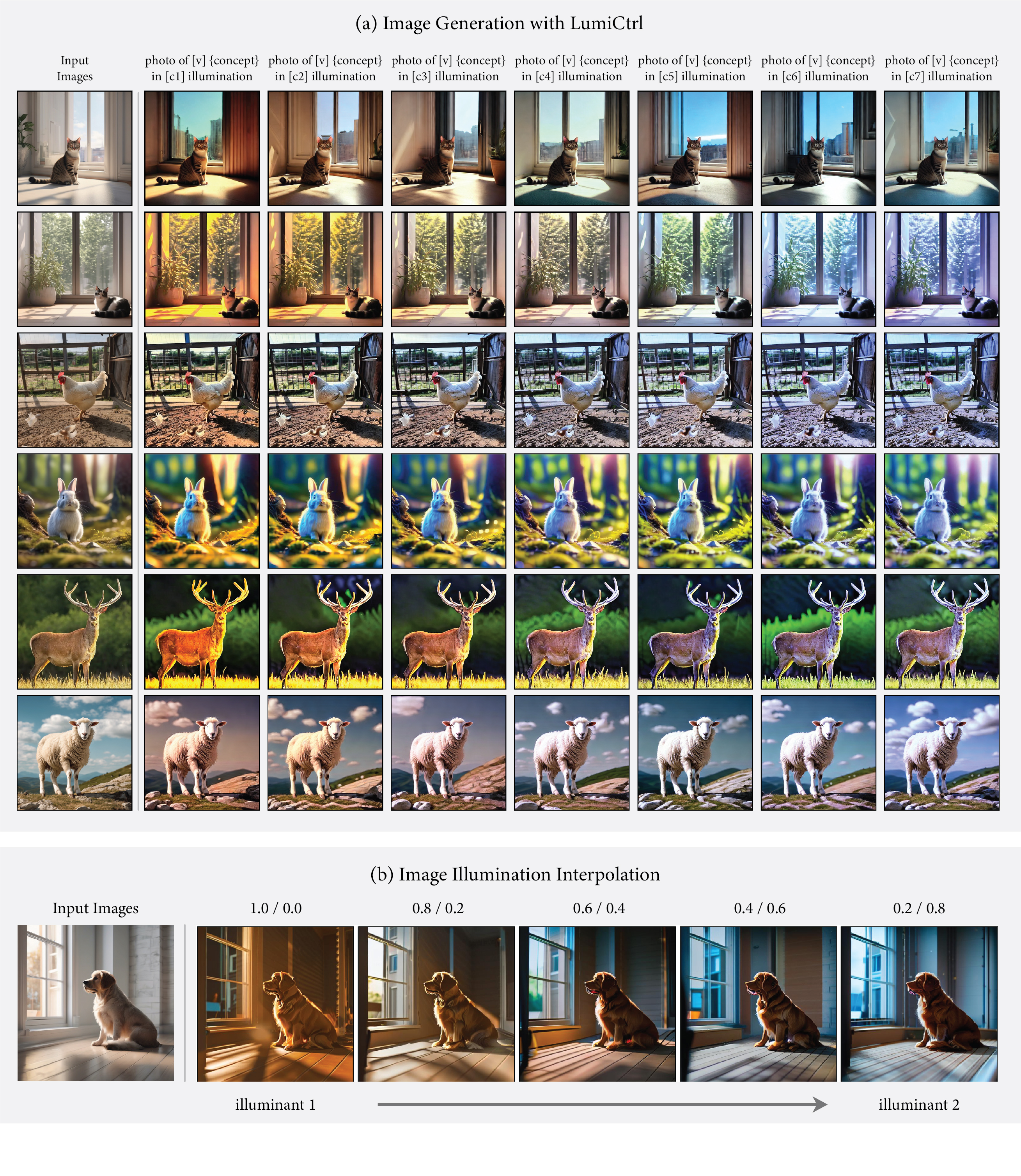}
\caption{Additional qualitative results of \ourmethod illuminating real and T2I generated concepts given text prompts. The \{concept\} in the prompt represents the name of the concept in the given images, such as cat, dog, rabbit, deer, and sheep.}
\label{fig:qual_res_supp}
\end{figure*}

Figure \ref{fig:illum_flat_ours} compares our method and a traditional illuminant adjustment based on Von Kries multiplication---\textit{Flat Light adaptation}. We can see how the \textit{contextual adaptation} presented in this paper generates a pleasant image. In contrast, the \textit{Flat Light adaptation} result is unrealistic, reducing the hue diversity in the scene (the top image becomes orangish, while the bottom image becomes bluish). In particular, we present a demonstration of the illuminating concepts under different illuminant conditions using the flat light adaptation in Fig.~\ref{fig:illum_fla}. It can be noted that the image becomes more unrealistic, i.e., extreme bluish or extreme orangish when the illuminant is scaled higher towards cool and warm conditions, respectively. Secondly, the flat light adaptation cannot create illuminance adaptive shadow, which is one of the main drawbacks of this approach. For instance, the shadow of the concepts is exactly same across all the illuminants which makes it unnatural, as the intensity of shadow of the concept scales with the changing daylight illumination in the real world. However, \ourmethod generates illuminance adaptive shadows as shown in Fig.~\ref{fig:qual_res_supp}.

\begin{figure*}[t]
\centering
\includegraphics[width=\linewidth]{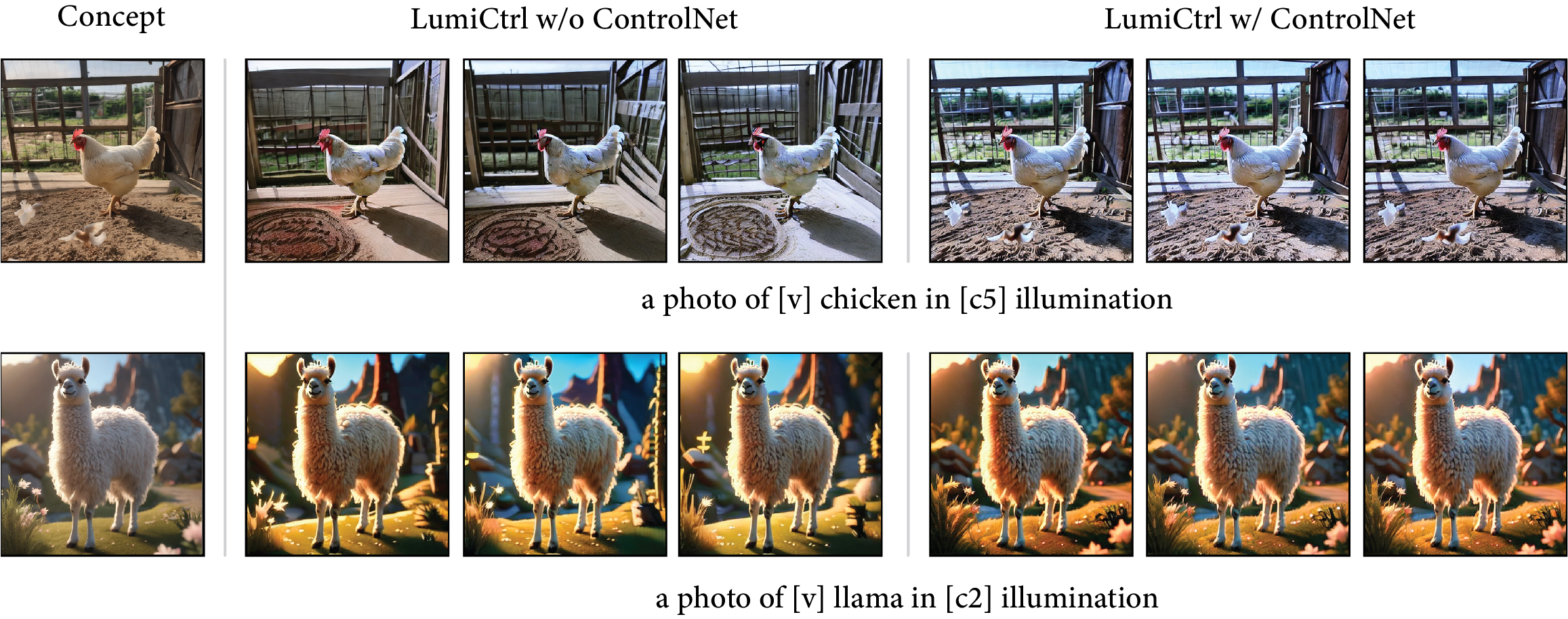}
\caption{Demonstrating the comparison between the \ourmethod's pipeline with and without Edge-Guided Prompt Disentanglement.}
\label{fig:ablate_lumictrl_supp}
\end{figure*}

\subsection{Qualitative Results}
We provide additional qualitative results demonstrating illumination of both---the real and T2I generated images into seven real-world daylight illuminations. The results are presented in Figure~\ref{fig:qual_res_supp} which shows the capabilities of \ourmethod in illuminating given concepts. We include outdoor and indoor examples to further analyze the versatility of \ourmethod. Though, \ourmethod currently provide 7 illuminants, however, we show in Figure~\ref{fig:qual_res_supp}b that the user can easily interpolate between the two learned illuminants. In this way, user can synthesize their concepts in numerous intermediate illuminants between the given seven illuminants. In addition, we also ablate the controlnet in \ourmethod's pipeline, and illustrated a comparitive qualitative analysis in Fig.~\ref{fig:ablate_lumictrl_supp}. The results show that \ourmethod struggles with preserving the structural information of the input image, and introduces artifacts in the generated images when controlnet based guidance is not integrated in the training pipeline. Whereas, this problem is mitigated, when controlnet based guidance is enabled in the \ourmethod pipeline. It is important to note that this conditioning mechanism is only used while learning the new concept. \ourmethod do not require ControlNet during inference to maintain generation quality.

\subsubsection{Ablation Study.}
We conduct ablation study over various factors. We note that \ourmethod introduces artifacts in generated images, when ControlNet based guidance is removed in the training, as shown in Fig.~\ref{fig:ablate_lumictrl_supp}. Moreover, \ourmethod generates unrealistic lighting and also affects background when $\lambda$ is scaled higher in masked reconstructed loss. We quantitatively analyze the effect of $\lambda$ in illumination quality in generated images. In this case, we show the results for the $4$ illuminants used in the user study. The results are summarized in Table~\ref{tab:ablation} which show that \ourmethod achieves better illumination quality with $\lambda=0.2$.
\vspace{20mm}
\begin{table}[!ht]
\centering
\small
\caption{Ablation study over the foreground weighting hyperparameter $\lambda$ in the Masked Reconstruction Loss. Lower Angular Error (AE) and MSE indicate better illuminant accuracy; higher SSIM indicate better image fidelity. Best performance is achieved at $\lambda = 0.2$.}
\label{tab:ablation}
\begin{tabular}{ccccc}
\hline
$\lambda$ & AE $\downarrow$ & MSE $\downarrow$ & SSIM $\uparrow$ \\
\hline
1.0 & 13.20 & 25.60 & 0.41 \\
0.8 & 10.10 & 22.30 & 0.58 \\
0.6 & 8.40  & 19.70 & 0.60 \\
0.4 & 6.90  & 18.10 & 0.65 \\
\textbf{0.2} & \textbf{4.51} & \textbf{16.80} & \textbf{0.77} \\
0.0 & 9.80  & 20.50 & 0.50 \\
\hline
\end{tabular}
\end{table}

\end{document}